\documentclass[journal]{IEEEtran}
\usepackage{cite}
\usepackage{amsmath,amssymb,amsfonts}
\usepackage{algorithmic}
\usepackage{graphicx}
\usepackage{booktabs}       
\usepackage{textcomp}
\usepackage{xcolor}
\usepackage{bm}
\usepackage{mathtools}
\usepackage{caption}
\usepackage{subcaption}

\def\BibTeX{{\rm B\kern-.05em{\sc i\kern-.025em b}\kern-.08em
    T\kern-.1667em\lower.7ex\hbox{E}\kern-.125emX}}
    
\definecolor{green1}{RGB}{50,175,120}

\ifCLASSINFOpdf
\else
\fi
\begin{document}
%
\title{Edge but not Least: Cross-View Graph Pooling}
%
%
%

\author{Xiaowei Zhou,
        Jie~Yin,~\IEEEmembership{Member,~IEEE}
        and Ivor~W.~Tsang,~\IEEEmembership{Senior Member,~IEEE}
\thanks{Preprint.}}

\maketitle

\begin{abstract}
Graph neural networks have emerged as a powerful model for graph representation learning to undertake graph-level prediction tasks. 
Various graph pooling methods have been developed to coarsen an input graph into a succinct graph-level representation through aggregating node embeddings obtained via graph convolution. However, most graph pooling methods are heavily node-centric and are unable to fully leverage the crucial information contained in global graph structure.
This paper presents a cross-view graph pooling (Co-Pooling) method to better exploit crucial graph structure information. The proposed Co-Pooling fuses pooled representations learnt from both node view and edge view. Through cross-view interaction, edge-view pooling and node-view pooling seamlessly reinforce each other to learn more informative graph-level representations. Co-Pooling has the advantage of handling various graphs with different types of node attributes. Extensive experiments on a total of 15 graph benchmark datasets validate the effectiveness of our proposed method, demonstrating its superior performance over state-of-the-art pooling methods on both graph classification and graph regression tasks.
\end{abstract}


%
\IEEEpeerreviewmaketitle

\section{Introduction}
%
%
%
%
\IEEEPARstart{G}{raph-structured} data is becoming ubiquitous across a wide variety of domains, such as chemical molecules~\cite{debnath1991structure}, social networks~\cite{yanardag2015deep,rozemberczki2020api}, financial networks~\cite{hamilton2020graph}, and citation networks~\cite{morris2020tudataset}. Learning effective graph representations plays a crucial role for many tasks across various application areas, such as drug discovery~\cite{jiang2021could}, molecule property prediction~\cite{li2017learning}, traffic forecast~\cite{jiang2021graph}, and so on. Recently, graph neural networks (GNN) have emerged as state-of-the-art models for graph representation learning, including 
graph convolutional network (GCN)\cite{kipf2017semi}, graph attention network (GAT)~\cite{velivckovic2018graph}, graph isomorphism network (GIN)~\cite{xu2019powerful}, and GraphSAGE~\cite{hamilton2017inductive}. The majority of these GNN models rely on message passing schemes in graph convolution to learn the embedding of each node by aggregating and transforming the embeddings of its neighbouring nodes. To obtain the representation of the entire graph, node embeddings are aggregated via a readout function or graph pooling methods~\cite{zhang2019hierarchical,ying2018hierarchical,ranjan2020asap,yuan2020structpool}. Graph pooling methods focus on coarsening an input graph into a compact vector-based representation for the entire graph, which is used for graph prediction tasks, such as graph classification or graph regression.




\begin{figure}[tbp]
    \centering
    \includegraphics[width=1.0\linewidth]{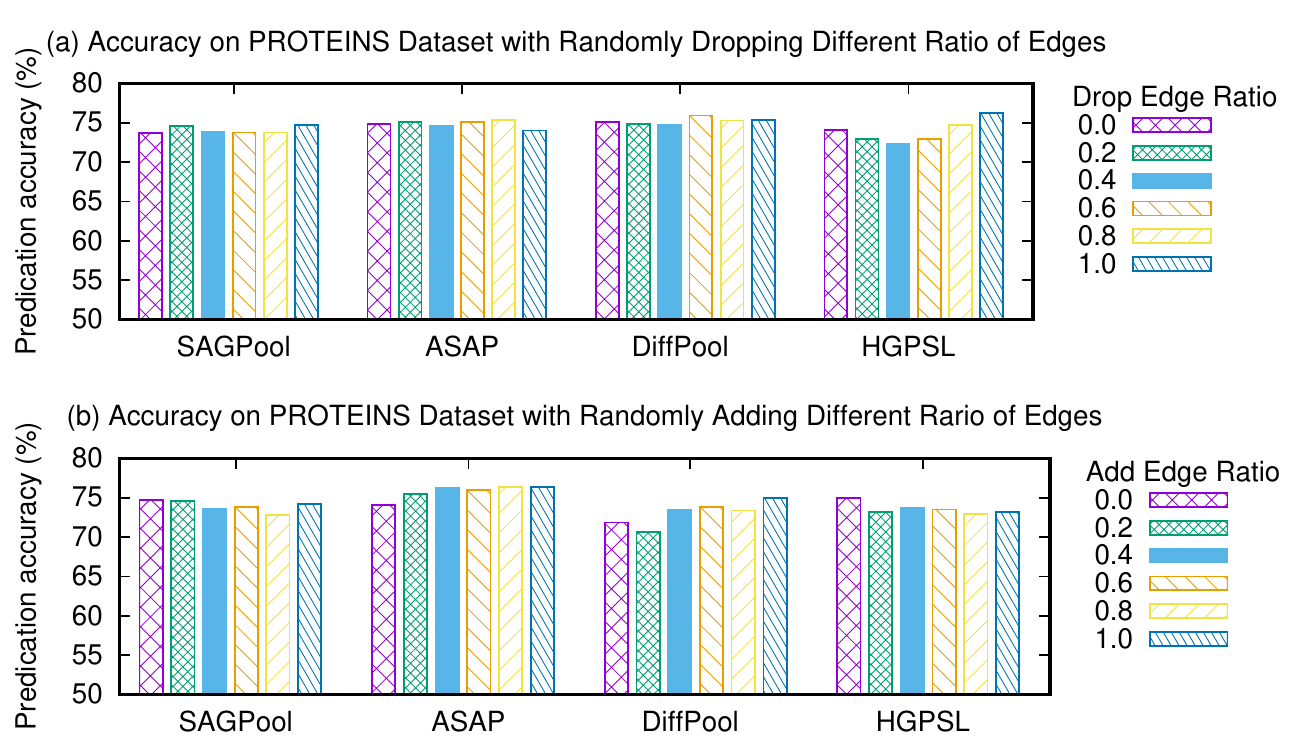}
    \caption{Classification accuracy on PROTEINS dataset with different edge ratios. (a) The classification accuracy does not significantly drop as we randomly drop different ratios of edges. (b) The classification accuracy does not significantly change when we randomly add different ratios of edges from no-edge graphs. Better view in colors.}
    \label{fig:noedge}
\end{figure}

To learn informative graph representations, a myriad of graph pooling methods have been proposed, which can be roughly categorized as node sampling based pooling and node clustering based pooling. Node sampling based pooling methods (e.g., SAGPool~\cite{lee2019self}, ASAP~\cite{ranjan2020asap}, HGPSL~\cite{zhang2019hierarchical}) typically calculate an importance score for each node and select the top $K$ important nodes to generate an induced subgraph. For example, SAGPool~\cite{lee2019self} selects nodes by learning importance scores via a self-attention mechanism. HGPSL~\cite{zhang2019hierarchical} samples the important nodes and uses an additional structure learning mechanism to learn new graph structure for sampled nodes. Node clustering based pooling methods, like differentiable graph pooling (DiffPool)~\cite{ying2018hierarchical}, learn an assignment matrix to cluster nodes into several super-nodes level by level. Through this process, a hierarchy of the induced subgraphs can be generated for representing the whole graph. However, we argue that the existing pooling methods focus primarily on aggregating node-level information for learning graph-level representations, but they fail to exploit the key graph structure. The loss of information present in the global graph structure would further hinder the message passing in subsequent layers. 



To verify our argument, we selected four state-of-the-art pooling methods: SAGPool~\cite{lee2019self}, ASAP~\cite{ranjan2020asap}, DiffPool~\cite{ying2018hierarchical}, and HGPSL~\cite{zhang2019hierarchical} and analyzed the influence of changing graph topological structure on the graph classification accuracy. We used the PROTEINS dataset as a case study, where we randomly dropped and added edges with different ratios. As shown in Fig.~\ref{fig:noedge}, we found that the random edge manipulation does not cause a significant decrease in the graph classification accuracy. Surprisingly, when there are no edges at all, i.e. dropping 100\% edges in Fig~\ref{fig:noedge}(a) and adding 0\% edges (no edges) in Fig~\ref{fig:noedge}(b), the classification accuracy still retains at the same level with other edge ratios. 
Especially for HGPSL that implicitly uses edge information, the classification accuracy is the highest when all edges are removed. Our empirical studies indicate the current graph pooling methods are heavily node-centric and are unable to fully leverage the crucial information contained in graph structure.

To fill this research gap, we propose a novel cross-view graph pooling method called Co-Pooling that explicitly exploits graph structure for learning graph-level representations. Our main motivations are twofold. First, we would like to capture crucial graph structure through explicitly pruning unimportant edges. Key structure information, such as functional groups (i.e., rings) in bio-molecular networks, or cliques in protein-protein interaction networks and social networks, has been widely recognised as a crucial source for graph prediction tasks~\cite{Milo2019network}. Second, real-world graphs may have various properties such as one-hot node attributes, real-valued node attributes, or even no attributes (see Table~\ref{tab:dataset}). Hence, we would like our new pooling method to seamlessly handle different types of graphs and make the best of node-level information when available.

Specifically, Co-Pooling is comprised of two key components: \textit{edge-view pooling} and \textit{node-view pooling}. The aim of edge-view pooling is to preserve crucial graph structure, which can benefit subsequent graph prediction tasks. This is achieved by capturing high-order structural information via generalized PageRank and pruning the edges with lower proximity weights. For node-view pooling, the importance score for each node is computed and the top important nodes are selected for pooling. The learning of graph pooling from edge and node views seamlessly reinforces each other through exchanging the proximity weights and the selected important nodes. The final pooled graph is obtained by fusing graph representations learnt from two views. Through cross-view interaction, Co-Pooling enables edge-view pooling and node-view pooling to complement each other towards learning effective graph representations. 



Our contributions are summarised as follows:
\begin{enumerate}
    \item We empirically analyze the ineffectiveness of the existing node-centric graph pooling methods in fully leveraging graph structure. 
    
    \item We propose a new cross-view graph pooling (Co-Pooling) method to learn graph representations by fusing the pooled graph information from both node view and edge view. The proposed method has the flexibility to handle different types of graphs (labeled/attributed graphs and plain graphs).
    \item We verify the effectiveness of Co-Pooling in graph classification and regression tasks across a wide range of 15 graph benchmark datasets, demonstrating its superiority over state-of-the-art pooling methods.
    
\end{enumerate}






\begin{figure*}[tbp]
    \centering
    \includegraphics[width=1.0\textwidth]{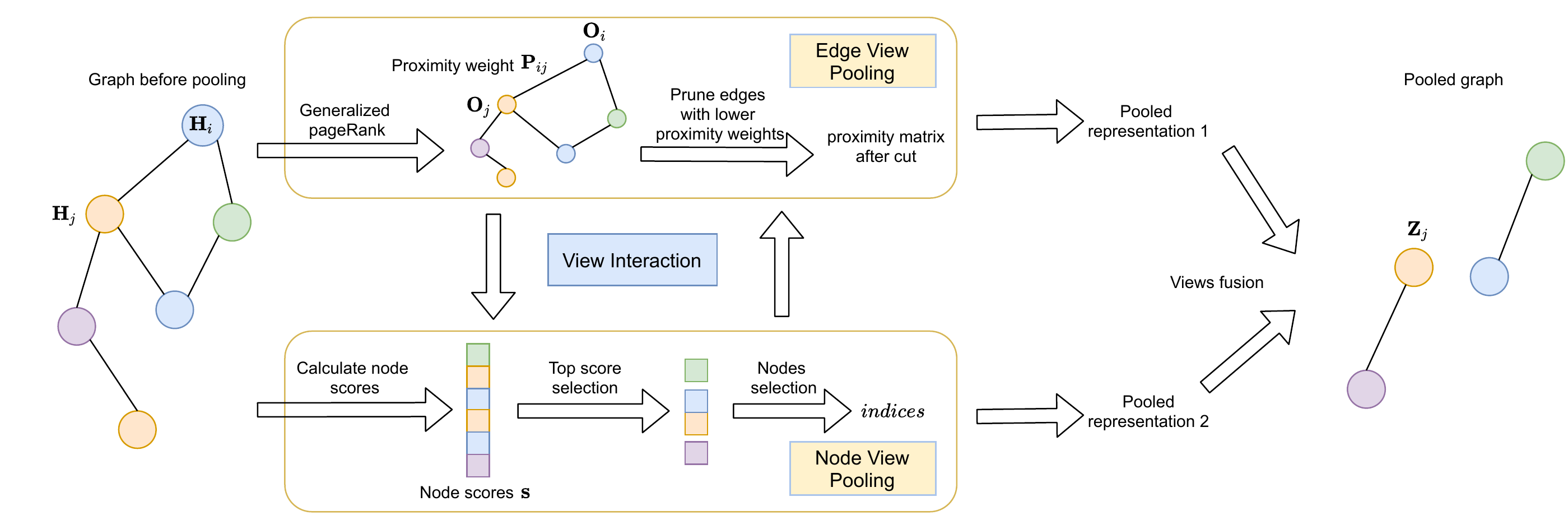}
    \caption{Overview of our proposed cross-view graph pooling. From the edge view, the higher-order structural information is captured via generalized PageRank and the proximity matrix is calculated to prune less important edges. From the node view, the top important nodes are selected for pooling. Edge-view pooling and node-view pooling seamlessly reinforce each other through view interaction. The pooled representations from both views are fused to form the final graph-level representation.}
    \label{fig:flowchat}
\end{figure*}



\section{Related Work}

Graph pooling is a key component in GNNs for learning a vector representation of the whole graph. The existing graph pooling methods can be divided into two categories: \textit{sampling based pooling} and \textit{clustering based pooling}. 

Sampling based pooling methods generate a smaller induced graph by selecting the top important nodes according to certain importance scores of nodes. There has been a series of graph pooling methods that fall into this category. Gao et al.~\cite{gao2019graph} proposed a pooling method that selects the top-$k$ nodes to form a smaller graph. This method projects node features into scalar values and uses them as the node selection criterion. Similarly, SAGPool~\cite{lee2019self} employs a self-attention mechanism to calculate the importance score for each node, and then chooses the top-ranked nodes to induce the pooled graph. Ranjan et al.~\cite{ranjan2020asap} proposed adaptive structure aware pooling (ASAP) for learning graph representations. This method updates node embeddings by aggregating features of neighbouring nodes in a local region. After that, a fitness score for the updated node is calculated to select the top-$k$ nodes to form the pooled graph. These methods, however, do not leverage graph structure in the pooling process. HGPSL~\cite{zhang2019hierarchical} takes a step forward to learn new edges between the top important nodes selected for the pooled graph. However, this method does not preserve the structure information contained in the original graph.


On the other hand, clustering based pooling methods learn an assignment matrix to cluster nodes into super-nodes. DiffPool~\cite{ying2018hierarchical} learns a differentiable soft cluster assignment for nodes in an end-to-end manner. The learnt assignment is used for grouping nodes in the last layer to several clusters in the subsequent layer. HaarPooling~\cite{wang2020haar} relies on compressive Haar transform filters to generate the induced graph of smaller size. 

Most of the existing pooling methods operate on a single node view; they are unable to fully leverage crucial graph structure. Although preliminary attempts (e.g., EdgePool~\cite{diehl2019edge} and EdgeCut~\cite{galland2021graph}) have been made to pool the input graph from an edge view, these methods simply rely on local connectivity to calculate pairwise edge scores and suffer from high computational complexity. In contrast, our edge-view pooling mechanism leverages higher-order structure information to assess the importance of edges, which is fed to further guide the selection of important nodes for node-view pooling. To the best of our knowledge, our work is the first to propose a cross-view graph pooling method, which enables us to fuse useful information from both edge and node views towards learning effective graph-level representations.

\section{Method}

This section presents the overview of our proposed cross-view graph pooling, followed by a detailed description of the main components.


\subsection{Preliminaries}
We first define notations to clarify the description of our proposed method. Assume we have $m$ input graphs $\mathbb{G}=\{G^{(0)}, G^{(1)}, \cdots,  G^{(m)}\}$ and their corresponding targets $\mathbb{Y}=\{y^{(0)}, y^{(1)}, \cdots, y^{(m)}\}$. For graph classification, $y_i$ is a discrete class label; for graph regression, $y_i$ is a continuous regression target variable $y_i\in \mathbb{R}$. A graph $G^{(g)}$ is represented as $(\mathbf{X}^{(g)}, \mathbf{A}^{(g)})$, where $\mathbf{X}^{(g)} \in \mathbb{R}^{n\times d}$ represents the node attribute matrix, where $n$ is the number of nodes and $d$ is the dimension of node attributes; $\mathbf{A}^{(g)}$ is the adjacent matrix, if there is an edge between node $i$ and node $j$, $\mathbf{A}^{g}_{ij}=1$; otherwise $\mathbf{A}^{g}_{ij}=0$ . For simplicity,  $(\mathbf{X}^{(g)}, \mathbf{A}^{(g)})$ is also noted as $(\mathbf{X}, \mathbf{A})$ to represent an arbitrary graph. $\hat{\mathbf{A}}=\mathbf{A}+\mathbf{I}$ stands for the adjacent matrix with self-loops.

In this paper, we use graph convolution network (GCN) as our backbone to learn representations for graphs. The graph convolution operation is defined as:

\begin{equation}
    \mathbf{H}=\hat{\mathbf{D}}^{-1/2}\hat{\mathbf{A}}\hat{\mathbf{D}}^{-1/2}\mathbf{X}\mathbf{\Theta}
\end{equation}
where $\mathbf{H} \in \mathbb{R}^{n\times f}$ is node embedding after convolution, $f$ is the dimension of node embedding; $\hat{\mathbf{D}}_{ii} = \sum_{j=0} \hat{\mathbf{A}}_{ij}$ is diagonal degree matrix; $\mathbf{\Theta}$ is a learnable parameter. After node embeddings $\mathbf{H}$ are learnt, graph pooling operation is applied to aggregate node embeddings to form a vector representation for the whole graph. This graph-level representation can be used for downstream graph prediction tasks, i.e. graph classification and graph regression.

\subsection{Cross-View Graph Pooling}
The proposed cross-view graph pooling (Co-Pooling) consists of two main components: edge-view pooling and node-view pooling, as illustrated in Fig.~\ref{fig:flowchat}, Co-Pooling simultaneously performs pooling from both edge view and node view. Through cross-view interaction, edge-view pooling and node-view pooling seamlessly reinforce each other, and finally the pooled graphs from two views are fused to form the final graph representation.  


\subsubsection{\textbf{Edge-View Pooling}}
The key objective of edge-view pooling is to preserve crucial information contained in graph structure. This is achieved by capturing high-order structural information via generalized PageRank (GPR)~\cite{chien2021adaptive} and pruning unimportant edges. 
Through edge-view pooling, the learnt representation captures a better connectivity relationship between nodes and higher-order graph structure information.

Specifically, we first update node embeddings by generalized PageRank to capture the information from higher-hop neighbours. As shown in Eq. (\ref{eq:gpr}), the node embeddings are updated by multiplying different GPR weights $\beta _t$. When $t=0$, we have $\mathbf{H}^0 = \mathbf{H}$; while $t > 0$, we have $\mathbf{H}^t = \hat{\mathbf{D}}^{-1/2}\hat{\mathbf{A}}\hat{\mathbf{D}}^{-1/2}\mathbf{H}^{t-1}$. Through generalized PageRank, node embeddings propagate $T$ steps. For each step, the GPR weight $\beta _t$ is learnable. Therefore, the contribution of each propagation step towards node embeddings can be learnt adaptively. The GPR operation of $T$ steps helps to incorporate higher-hop neighbours' information to update node embeddings. 

\begin{equation}
    \mathbf{O}=\sum_{t=0}^T\beta_{t}\mathbf{H}^{t}
    \label{eq:gpr}
\end{equation}

After updating node embeddings by generalized PageRank, we calculate the proximity weights between each pair of nodes, and the edges with low proximity weights can be pruned to only preserve crucial graph structure. 

The process of computing the proximity weights can be illustrated using 
Eq.~(\ref{eq:att}), where $\mathbf{O}_i$ and $\mathbf{O}_j$ are the GPR updated embeddings of node $i$ and node $j$. We first transform node embeddings $\mathbf{O}_i$ and $\mathbf{O}_j$ via a linear transformation parameterized with $\mathbf{W}$ and then concatenate the transformed embeddings. Another linear transformation with learnable parameters $\mathbf{a}$ is used to transform the concatenated embeddings. Finally, the proximity weight $\mathbf{P}_{ij}$ between node $i$ and node $j$ is obtained after a Sigmoid function. 

\begin{equation}
    \mathbf{P}_{ij}=\sigma(\mathbf{a}^{T}[\mathbf{W}\mathbf{O}_i\lVert \mathbf{W}\mathbf{O}_j]) * \mathbf{A}_{ij},
    \label{eq:att}
\end{equation}
where $\mathbf{P} _{ij}$ is the proximity weight between node $i$ and node $j$; $\sigma$ is Sigmoid function; $\lVert$ represents the concatenation operation; $\mathbf{a}$ and $\mathbf{W}$ are learnable parameters; $\mathbf{A}_{ij}=1\ or\ 0$ indicates whether or not there is an edge connecting node $i$ and node $j$. 

According to the proximity weight $\mathbf{P} _{ij}$ of each node pair, we can obtain the proximity matrix $\mathbf{P}$ for all node pairs in a graph. To emphasize the proximity between each node and itself, we update the proximity matrix by adding value 1 in diagonal, i.e., $\mathbf{\hat{P}}=\mathbf{P} + \mathbf{I}$. For undirected graphs, we average the proximity weights at symmetric positions by $\mathbf{P}_{sym}=(\mathbf{\hat{P}} + \mathbf{\hat{P}}^T)/2$. 

Based on the proximity matrix $\mathbf{P}_{sym}$, we prune unimportant edges with low proximity weights in the graph. For a given edge retaining ratio $\gamma$, we have the cut proximity matrix $\mathbf{P}_{cut}=\textrm{Top}_{\gamma}(\mathbf{P}_{sym})$, where $\textrm{Top}_{\gamma}()$ is the operation that retains the top $\gamma$ percentage of edges with high proximity weights. Accordingly, we update the adjacent matrix to reflect the removal of edges. The cut proximity matrix provides a better measure to quantify the higher-order connectivity relationship between nodes, which is fed to node-view pooling to guide the selection of important nodes. 


\subsubsection{\textbf{Node-View Pooling}}

For node-view pooling, the aim is to select the top $K$ important nodes for coarsening the input graph. To better measure the connectivity between nodes, we take the cut proximity matrix from edge-view pooling to compute an importance score for each node, given by
\begin{equation}
    \label{eq:score}
    {\mathbf{s}}=\hat{\mathbf{D}_{cut}}^{-1/2}\hat{\mathbf{P}_{cut}}\hat{\mathbf{D}_{cut}}^{-1/2}\mathbf{H} {\mathbf{1}}^T
\end{equation}
where $\mathbf{s}$ is the score vector for all nodes, $\hat{\mathbf{D}_{cut}}$ is diagonal degree matrix of $\hat{\mathbf{P}_{cut}}$, $\hat{\mathbf{P}_{cut}}=\mathbf{P}_{cut}+\mathbf{I}$, $\mathbf{1}^T$ is the vector containing all ones. 

Based on nodes' importance scores, we select the top $K = n * \epsilon$ nodes, where $\epsilon$ is the node pooling ratio. After selecting the nodes, we obtain the $indices$ and embeddings of all selected nodes. 


\subsubsection{\textbf{Edge-Node View Interaction}}

To enable edge-view pooling and node-view pooling to reinforce each other, our Co-Pooling method exchanges the cut proximity matrix and the indices of selected nodes, which serve as the mediator for the interaction between two views.

For node-view pooling, the cut proximity matrix from edge-view pooling is used to calculate the important score for each node. The cut proximity matrix better reflects higher-order connectivity relationship between nodes, thus providing more accurate information than the original adjacent matrix to quantify the importance of nodes. After obtaining the node scores, we select the top-K important nodes as the pooled representation from node-view pooling, i.e. $\mathbf{H}(indices, :)$, where $(indices, :)$ represents index selection operation.

For edge-view pooling, the indices of selected nodes obtained from node-view pooling are used to cluster nodes into super-nodes. The important node indices in node-view pooling are useful to guide the clustering operation, as the selected important nodes are determined by considering higher-order structure information. The pooled representation from edge-view pooling is obtained through $\mathbf{P}_{cut}(indices,:) \times \mathbf{O}$, where $\times$ means matrix multiplication. 

Lastly, the pooled representations from node-view pooling and edge-view pooling are fused to form the final graph representation as:
\begin{equation}
    \mathbf{Z}=\mathbf{W}[\mathbf{P}_{cut}(indices,:)\times\mathbf{O} \lVert \mathbf{H}(indices,:)]
    \label{eq:fuse}
\end{equation}
where $\mathbf{W}$ is the learnable parameter for linear transformation; $\lVert$ represents the concatenation operator. $\mathbf{Z}$ is the graph-level representation after pooling. Through edge-node view interaction, our Co-Pooling method enables edge-view pooling and node-view pooling to complement each other towards learning more effective graph representations. 

\begin{table*}[tbp]
    \centering
    \caption{Details of graph datasets for graph classification evaluation.}
    \begin{tabular}{c|c|c|c|c|c|c}
    \toprule
    Data set & \# of Graphs & \# of Classes & Avg. \# of Nodes & Avg. \# of Edges & Node Attributes & Dataset Type \\
    \midrule
    BZR-A & 405 & 2 & 35.75 & 38.36 & Real-valued Attribute & \textit{Attributed} \\
    AIDS-A & 2000 & 2 & 15.69 & 16.20 & Real-valued Attribute & \textit{Attributed} \\
    FRANKENSTEIN & 4337 & 2 & 16.90 & 17.88 & Real-valued Attribute & \textit{Attributed} \\
    \midrule
    PROTEINS & 1113 & 2 & 39.06 & 72.82 & Node label & \textit{Labeled} \\
    D\&D & 1178 & 2 & 284.32 & 715.66 & Node label & \textit{Labeled} \\
    NCI1 & 4110 & 2 & 29.87 & 32.30 & Node label  & \textit{Labeled} \\
    NCI109 & 4127 & 2 & 29.68 & 32.13 & Node label  & \textit{Labeled} \\
    MSRC\_21 & 563 & 20 & 77.52 & 198.32 & Node label  & \textit{Labeled} \\
    \midrule
    COLLAB & 5000	& 3	& 74.49	& 2457.78 & None & \textit{Plain} \\
    IMDB-BINARY & 1000 & 2 & 19.77 & 96.53 & None & \textit{Plain} \\
    IMDB-MULTI & 1500 & 3 & 13.00 & 65.94 & None & \textit{Plain} \\
    REDDIT-BINARY & 2000 & 2 & 429.63 & 497.75  & None & \textit{Plain}  \\
    REDDIT-MULTI-12K & 11929 & 11 & 391.41 & 456.89 & None & \textit{Plain} \\
    \bottomrule
    \end{tabular}
    \label{tab:dataset}
\end{table*}

\section{Experiments}
\begin{figure}[htbp]
    \centering
    \includegraphics[width=1.0\linewidth]{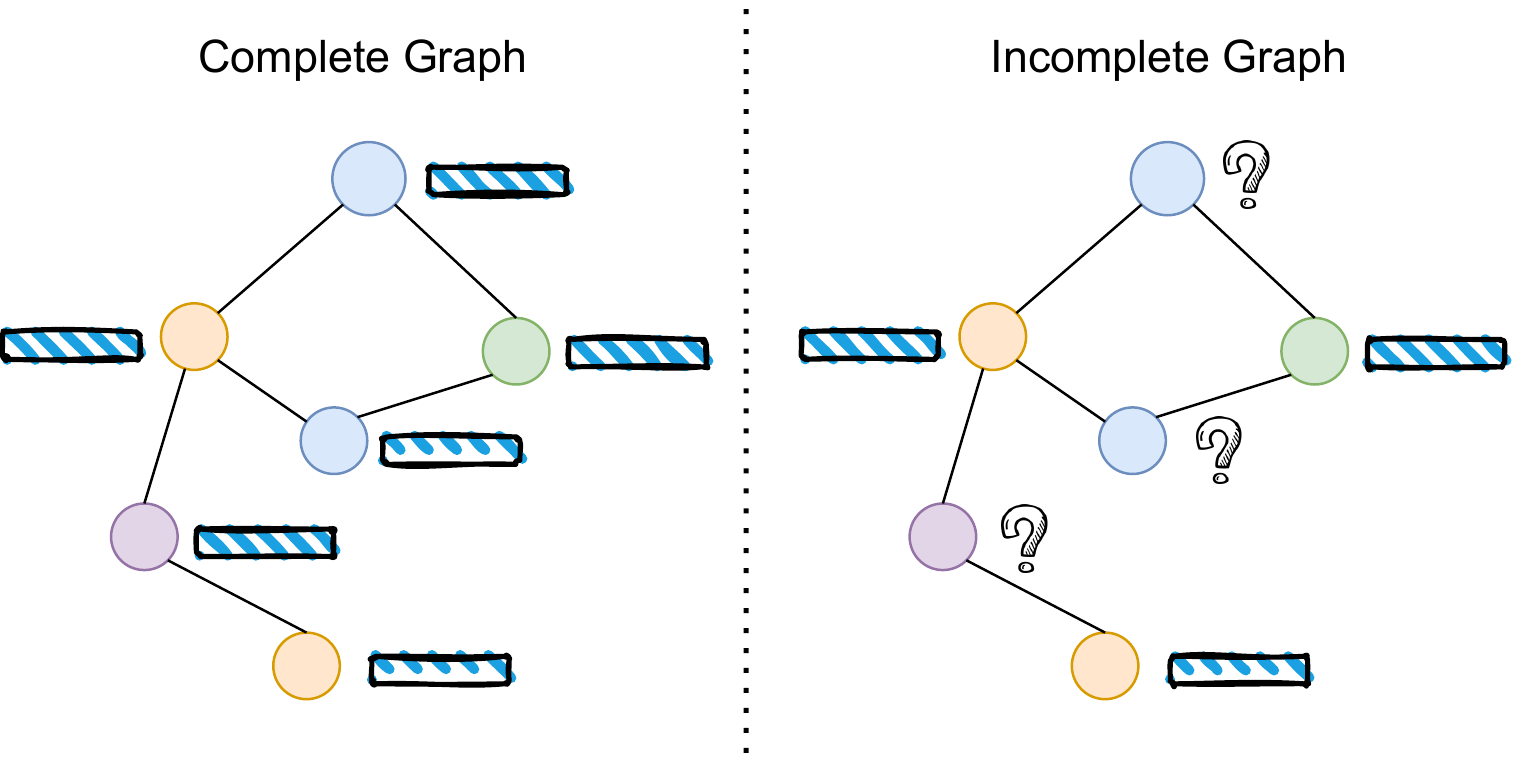}
    \caption{Comparison between complete graphs and incomplete graphs.}
    \label{fig:incompGraph}
\end{figure}

In this section, we validate the performance of our proposed cross-view graph pooling method on both graph classification and graph regression tasks. For graph classification task, we compare our method with several state-of-the-art pooling methods at two settings: complete graphs with various types of node attributes and incomplete graphs. As illustrated in Fig.~\ref{fig:incompGraph}, complete graphs refer to the graphs with all node attributes, while incomplete graphs are the ones with a portion of nodes having completely missing attributes. The incomplete graph setting is used to simulate real-world scenarios where attribute information for some nodes are inaccessible due to privacy or legal constraints. Our method is also compared against baseline pooling methods on graph regression task. 


%

\subsection{Graph Classification on Complete Graphs}
\label{ComGraphClassification}

\paragraph{\textbf{Benchmark Datasets}}
We conduct graph classification tasks on a total of 13 benchmark graph datasets with various attribute properties, including three \textit{attributed graphs} with real-valued node attributes, five \textit{labeled graphs} with only one-hot node attributes, and five \textit{plain graphs} with no node attributes. The detailed statistics about these datasets are listed in Table~\ref{tab:dataset}.


\begin{itemize}
\item \textbf{BZR-A}~\cite{sutherland2003spline} is a dataset of chemical compounds for classifying biological activities into active and inactive. The node attributes are 3D coordinates of the compound structures. 

\item \textbf{AIDS-A}~\cite{riesen2008iam} contains graphs representing molecular compounds. It contains two classes of graphs, which are against HIV or not. 

\item \textbf{FRANKENSTEIN}~\cite{orsini2015graph} consists of molecules as mutagens and non-mutagens for mutagenicity classification task. Node attributes are 780 dimensions MNIST~\cite{LecunMnist} image vectors of pixel intensities, which represent chemical atom symbols. 

\item \textbf{D\&D}~\cite{dobson2003distinguishing} and \textbf{PROTEINS}~\cite{borgwardt2005protein} include macromolecules as graph datasets in bioinformatics, which are for enzyme and non-enzyme classification task. 

\item \textbf{NCI1}~\cite{wale2008comparison} and \textbf{NCI109}~\cite{wale2008comparison} contain chemical compounds as small molecules, which are used for anticancer activity classification. 

\item \textbf{MSRC\_21}~\cite{neumann2016propagation} is graph dataset constructed by semantic images. Each semantic image is represented as a conditional Markov random field graph. Nodes in a graph represent the segmented superpixels in an image. If the segmented superpixels are adjacent, the corresponding nodes are connected. Each node is assigned with a semantic label as the node attribute. 

\item \textbf{COLLAB}~\cite{yanardag2015deep} is a scientific collaboration dataset, where each graph represents the collaboration network of one researcher. The task of COLLAB is to classify the graph into different research fields. 

\item \textbf{IMDB-BINARY}~\cite{yanardag2015deep} and \textbf{IMDB-MULTI}~\cite{yanardag2015deep} are two datasets for classifying each graph into different movie genres. Each graph is an ego-network for each actor/actress, and nodes also represent actors/actresses. 

\item \textbf{REDDIT-BINARY}~\cite{yanardag2015deep} and \textbf{REDDIT-MULTI-12k}~\cite{yanardag2015deep} are two datasets generated from online discussions. Each graph represents a discussion thread where nodes are different users. If one of two users responds to the other one, there is an edge between these two. The task is to classify which section the discussion belongs to.  
\end{itemize}

\begin{table*}[htbp]
\caption{Graph classification accuracy on benchmark graph datasets by different graph pooling methods. The best results are marked in bold. Higher is better. (- means the results can not be got in an acceptable time, i.e. 24h)}
 \label{tab:graphAcc}
 \centering
\begin{tabular}{c|c|c|c|c|c||c|c|c}
\toprule
Datasets & SAGPool & ASAP & DiffPool & HGPSL & EdgePool & Co-Pooling/GPR & Co-Pooling/NV & Co-Pooling \\
\midrule
BZR-A & 82.95$\pm$4.91 & 83.70$\pm$6.00 & 83.93$\pm$4.41 & 83.23$\pm$6.51 & 83.43$\pm$6.00 & 81.00$\pm$5.82 & 81.69$\pm$5.80 & \textbf{85.67$\pm$5.29} \\
AIDS-A & 98.85$\pm$0.78 & 99.00$\pm$0.74 & 99.40$\pm$0.58 & 99.10$\pm$0.66 & 99.05$\pm$0.69 &  98.85$\pm$0.71 &  98.90$\pm$0.58 &  \textbf{99.45$\pm$0.42} \\
FRANKENSTEIN & 60.94$\pm$2.90 & 66.73$\pm$2.76 & 65.08$\pm$1.50 & 62.19$\pm$1.74 & 62.99$\pm$2.21 & 64.01$\pm$1.70 & \textbf{67.00$\pm$2.37} & 64.15$\pm$1.34 \\
\midrule
D\&D & 76.91$\pm$3.42 & 77.84$\pm$3.41 & \textbf{78.01$\pm$2.70}  & 77.33$\pm$4.22 & 76.66$\pm$2.05 & 75.81$\pm$3.81 & 77.00$\pm$5.04 & 77.85$\pm$2.21 \\
PROTEINS & 73.68$\pm$4.63 & 74.85$\pm$5.18& 75.11$\pm$2.95  & 74.13$\pm$4.12 & \textbf{77.01$\pm$5.41} & 73.68$\pm$2.33 & 76.28$\pm$5.09    & 76.19$\pm$4.13    \\
NCI1 & 71.51$\pm$4.51 & 76.59$\pm$1.71& 74.14$\pm$1.43  & 73.48$\pm$2.42 & 78.39$\pm$2.43 & 77.25$\pm$2.11 & \textbf{79.15$\pm$2.04} & 78.66$\pm$1.48    \\
NCI109   & 69.69$\pm$3.27 & 74.73$\pm$3.48& 72.04$\pm$1.43 & 72.30$\pm$2.18 & 77.01$\pm$2.39 & 75.60$\pm$1.46 & \textbf{78.07$\pm$1.77} & 77.08$\pm$2.03    \\
MSRC\_21 & 90.22$\pm$2.82 & 90.41$\pm$3.91& 90.41$\pm$3.58  & 88.97$\pm$4.78 & 90.05$\pm$3.02 & 91.64$\pm$2.79 & 91.29$\pm$3.70 & \textbf{92.54$\pm$2.63} \\
\midrule
COLLAB & 70.58$\pm$2.31 & 72.84$\pm$1.84& 72.18$\pm$1.68  & 74.2$\pm$2.72 & - & 74.82$\pm$2.10 & 68.9$\pm$5.59     & \textbf{77.30$\pm$2.29} \\
IMDB-BINARY & 60.9$\pm$2.34  & 65.5$\pm$2.80 & 58.27$\pm$5.92  & 62.5$\pm$3.5 & 60.3$\pm$5.08 & 70.4$\pm$3.85  & 70.8$\pm$3.6& \textbf{72.1$\pm$4.44}  \\
IMDB-MULTI & 39.8$\pm$3.39  & 45.93$\pm$4.03& 40.00$\pm$4.52  & 40.53$\pm$4.88 & 44.27$\pm$4.50  & 47.6$\pm$4.55  & 44.8$\pm$3.94 & \textbf{49.07$\pm$3.28} \\
REDDIT-BINARY & 83.55$\pm$4.53 & - & 84.61$\pm$2.42  & - & 88.35$\pm$2.31 & 88.90$\pm$2.00 & 88.0$\pm$4.69 & \textbf{89.35$\pm$1.25} \\
REDDIT-MULTI-12K & 40.56$\pm$3.30 & - & 41.21$\pm$1.96 & - & - & 46.84$\pm$2.26 & \textbf{49.02$\pm$1.56} & {46.85$\pm$2.62} \\ 
\bottomrule
\end{tabular}
\end{table*}

\paragraph{\textbf{Baselines}}
We use five state-of-the-art graph pooling methods as our baselines: SAGPool~\cite{lee2019self}, ASAP~\cite{ranjan2020asap}, DiffPool~\cite{ying2018hierarchical}, HGPSL~\cite{zhang2019hierarchical}, and EdgePool~\cite{diehl2019edge}. When training DiffPool, we use the auxiliary link prediction loss function and entropy regularization items as did in the original paper. In addition, we also compare with two ablated variants of our cross-view graph pooling (Co-Pooling): Co-Pooling/GPR, and Co-Pooling/NV. Cross-pool/GPR is our cross-view graph pooling without generalized PageRank, and Co-Pooling/NV is our cross-view graph pooling without node-view pooling.

\paragraph{\textbf{Model Architecture and Training}}
In our experiments, the GNN used is built on the GCN architecture. The whole GNN consists of three GCN layers, two pooling layers and three linear transformation layers. After the last linear transformation layer, the SoftMax classifier is connected. Note that, the input to the first linear transformation layer is the concatenated features after each pooling layer. For all datasets, we use the same GNN architecture for fair comparison. The detailed architecture is provided in Appendix~\ref{app:GNN}.

When training the GNN model, we perform 10-fold cross-validation as did in~\cite{ying2018hierarchical}. We randomly split the dataset into training, validation, and test sets with 80\%, 10\% and 10\% graphs. We use Adam~\cite{kingma2015adam} optimizer for training the GNN model. The optimization stops if the validation loss does not improve after 50 epochs. The maximum epoch is set as 300. Following the strategy of searching optimal hyperparameters in ~\cite{lee2019self}, we use grid search to obtain optimal hyperparameters for each method. The ranges of different hyperparameters are as follows: learning rate in \{0.005, 0.0005, 0.001\}; weight decay in \{0.0001 0.001\}; node pooling ratio in \{0.5, 0.25\}; hidden size in \{128, 64\}; dropout ratio in \{0, 0.5\}.  To implement convolution operation on plain graph datasets where nodes have no attributes, we follow the implementation in DiffPool~\cite{ying2018hierarchical} to pad each node with a constant vector, i.e. all-one vector in a fixed dimension.

\paragraph{\textbf{Classification with State-of-the-art}}
We compare graph classification accuracy of all methods averaged over 10-fold cross-validation on each dataset. For a fair comparison, all baseline methods and our method are trained using the same training strategy. The GNN model architecture used for each method is also the same. As shown in Table~\ref{tab:graphAcc}, our cross-view graph pooling method achieve the best results across 11 datasets and the second place on the other two datasets. Particularly, our method significantly improves the best baseline method by 6.6\%, 3.14\%, 7.81\%, 2.13\% and 1.74\% on IMDB-BINARY, IMDB-MULTI, REDDIT-MULTI-12K, MSRC\_21 and BZR-A, respectively. This proves the effectiveness of our proposed method in predicting different types of graphs with various attribute properties. It is worth noting that our proposed cross-view graph pooling method achieves the best performance on all five datasets without node attributes. This shows the superiority of our method to complement the node-view pooling by edge-view pooling, when node attributes are not informative. 

When comparing different variants of our method, Co-Pooling consistently outperforms Co-Pooling/GPR on all datasets. This shows the importance of using generalized PageRank to capture the higher-order structure information. Co-Pooling yields higher accuracy than Co-Pooling/NV on most (8/13) of the datasets. This demonstrates the effectiveness of our method in combining two complementary views. On Labeled graphs, the performance of Co-Pooling and Co-Pooling/NV is comparable. This is because the important node indices used for clustering nodes in edge view may be inaccurate, as one-hot attributes provide limited information. On attributed graphs with real-valued node attributes and plain graphs with padding all-one vectors as node attributes, the obtained important node indices are more accurate to complement edge-view. 


\subsection{Graph Classification on Incomplete Graphs}
We also compare the performance of our method and baseline methods on incomplete graphs. For incomplete graphs, a portion of nodes have completely missing attributes, as shown in Fig.~\ref{fig:incompGraph}. This set of experiments is used to evaluate the effectiveness of our method on real-world scenarios, where attribute information for some nodes are inaccessible due to privacy or legal constraints.  

\paragraph{\textbf{Experimental Setup and Training}} We perform experiments on  \textit{attributed graph} AIDS-A and \textit{labeled graph} MSRC\_21 as a case study. For the two datasets, we randomly select different ratios of nodes and remove their original node attributes, while keeping all other remaining nodes unchanged. We define the ratio of nodes with all their attributes removed as incomplete ratio. For example, if we remove all attributes for 10\% of nodes, the incomplete ratio is 10\%. The resulting incomplete graph datasets are randomly divided into training set (80\%), validation set (10\%) and test set (10\%). We train the GNN model with different pooling methods on training set. The GNN model architecture used in this part is the same as that in Section~\ref{ComGraphClassification}. The best hyperparameters obtained from Section~\ref{ComGraphClassification} are used for training the GNN model in this part. Model architecture and training strategy for each method remain the same. We report graph classification accuracy averaged over 10-fold cross-validation. 

\begin{figure*}[htbp]
    \centering
    \includegraphics[width=1.0\textwidth]{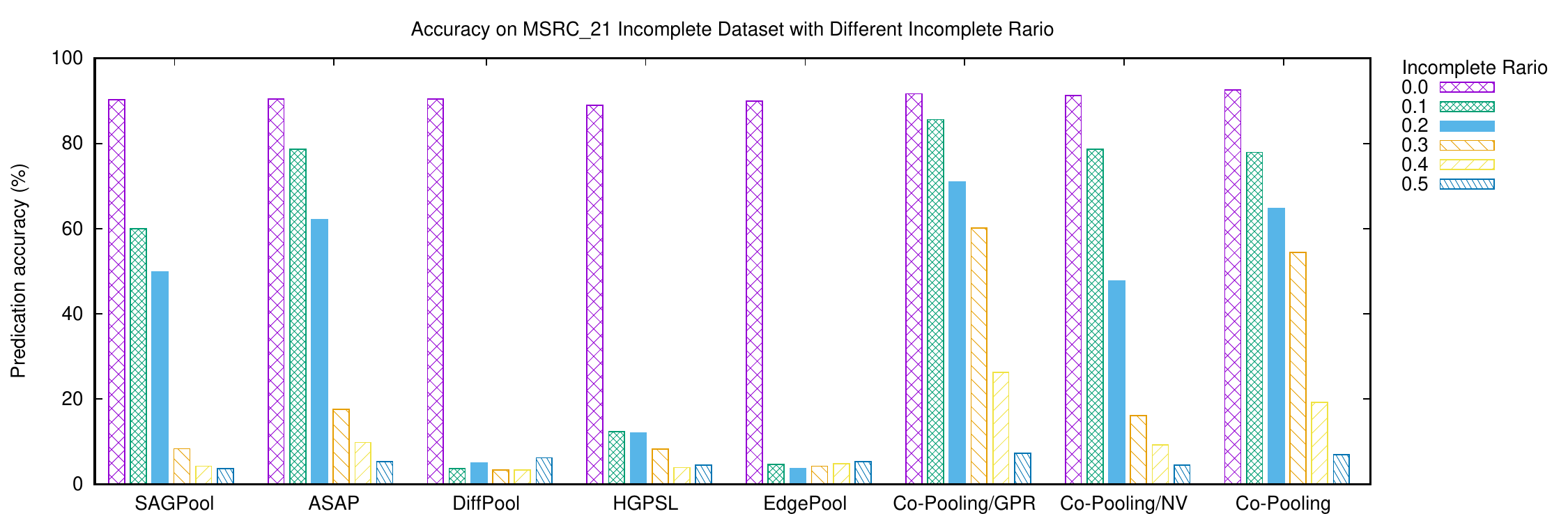}
    \caption{Graph classification accuracy on incomplete \textit{Labeled} graph dataset (MSRC\_21) by different graph pooling methods. Better view in colors.}
    \label{fig:NodeIncomMSRC21-100}
\end{figure*}

\paragraph{\textbf{Comparison. with State-of-the-art}}
Fig.~\ref{fig:NodeIncomMSRC21-100} compares the classification accuracy of all methods on MSRC\_21 incomplete dataset. For all baseline methods, the classification accuracy drops markedly as the incomplete ratio increases from 0\% to 50\%. For our Co-Pooling method, the accuracy decreases at a much slower rate than baseline methods. Especially for DiffPool, HGPSL and EdgePool, the classification accuracy drops by 3.73\%, 12.33\% and 4.61\%, respectively, even though only 10\% nodes have their attributes missing. Under the 10\% incomplete ratio, Co-Pooling and its variants can still achieve at least 77.93\% accuracy. Compared with the best baseline method ASAP, Co-Pooling achieves an average of 8.62\% increase in classification on all incomplete graph datasets with an incomplete ratio from 0\% to 50\%.

\begin{figure*}[htbp]
    \centering
    \includegraphics[width=1.0\textwidth]{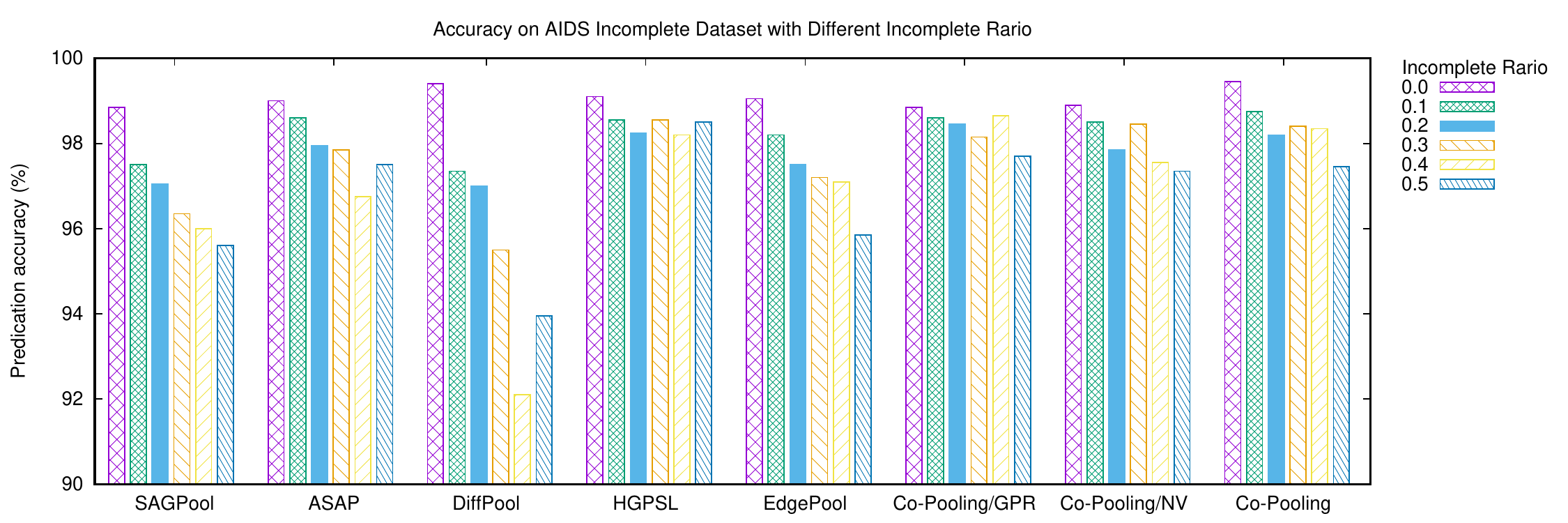}
    \caption{Graph classification accuracy on incomplete attributed graph dataset (AIDS-A) by different graph pooling methods. Better view in colors.}
    \label{fig:NodeIncomAIDS}
\end{figure*}

Fig.~\ref{fig:NodeIncomAIDS} compares graph classification accuracy of all methods on AIDS-A incomplete datasets. We can see that, for SAGPool, DiffPool and EdgePool, the classification accuracy decreases by 3.25\%, 5.45\%, and 3.2\%, respectively, when the incomplete ratio increases from 0\% to 50\%. In contrast, the accuracy of our methods drops only by 1.15\% under the same setting. Compared with ASAP, our methods can achieve better performance under all incomplete ratios. 
Compared with HGPSL, our methods achieve better performance on 0\%, 10\%, 20\% and 40\% incomplete graph datasets. Overall, our method still outperform HGPSL in terms of the average performance on all incomplete graph datasets,  

The classification comparisons on incomplete graph datasets demonstrate the effectiveness of our method in handling graphs with missing node attributes. This further testifies the complementary advantage of our method by fusing node-view and edge-view pooling, especially when node attributes are less informative. 


\subsection{Parameter Sensitivity}
\label{ablation}
The Co-Pooling method has the edge retaining ratio ($\gamma$) as an important parameter to determine how many percentage of edges are retained during edge-view pooling. To investigate the effect of the edge retaining ratio (i.e., $\gamma$) on the graph classification accuracy of Co-Pooling, we conduct empirical studies on six representative graph datasets, including two labeled graphs, two attributed graphs, and two plain graphs. On each graph dataset, we train the GNN model with the keeping ratio ranging from 10\% to 100\%. All other hyperparameters are set as the best parameters obtained in Section~\ref{ComGraphClassification}. We also use the same GNN model architecture and training strategy as in Section~\ref{ComGraphClassification}. We report the average classification accuracy on 10-fold cross-validation.

\begin{figure}
    \centering
    \begin{subfigure}[h]{0.49\columnwidth}
    \centering
    \includegraphics[width=\columnwidth]{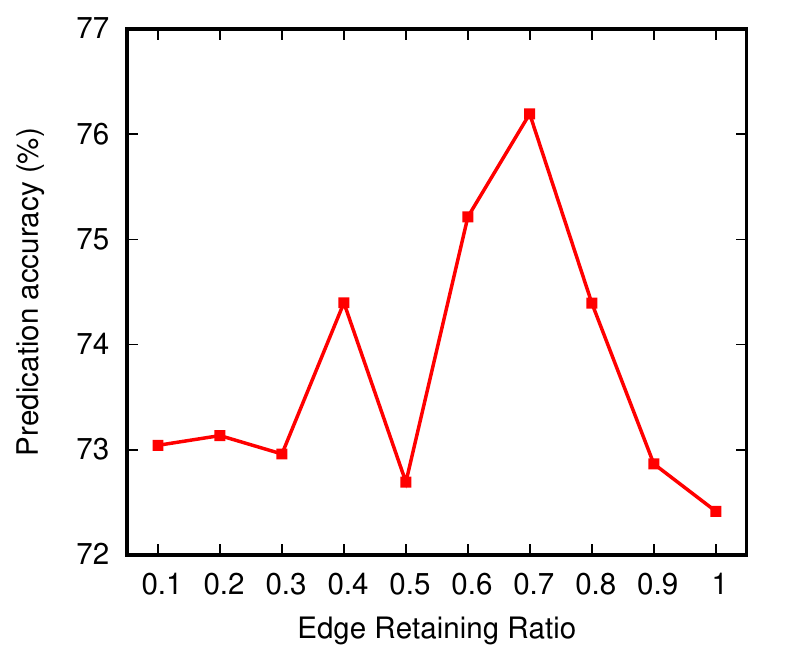}
    \caption{PROTEINS}
    \end{subfigure}
    \hfill
    \begin{subfigure}[h]{0.49\columnwidth}
    \centering
    \includegraphics[width=\columnwidth]{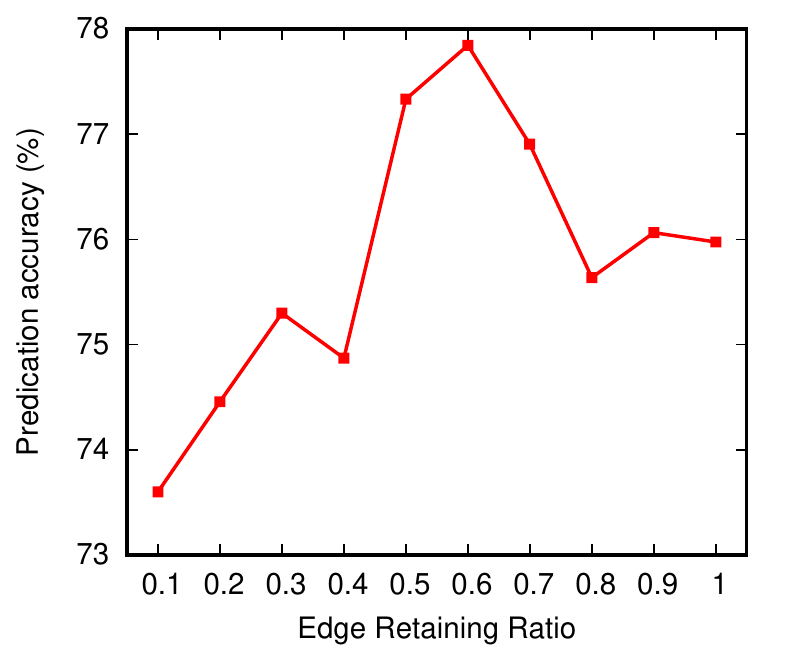}
    \caption{D\&D}
    \end{subfigure}
    
    \begin{subfigure}[h]{0.49\columnwidth}
    \centering
    \includegraphics[width=\columnwidth]{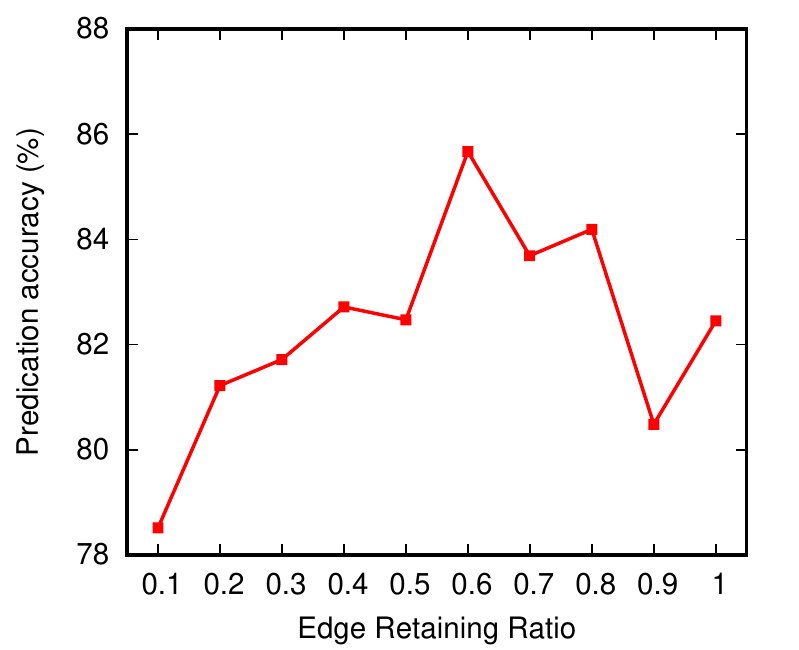}
    \caption{BZR-A}
    \end{subfigure}
    \hfill
    \begin{subfigure}[h]{0.49\columnwidth}
    \centering
    \includegraphics[width=\columnwidth]{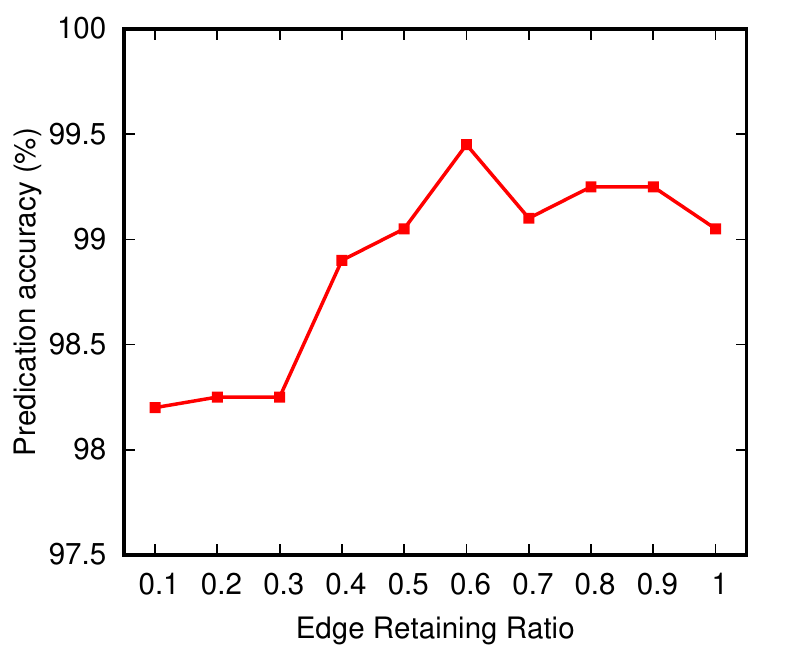}
    \caption{AIDS-A}
    \end{subfigure}
    
    \begin{subfigure}[h]{0.49\columnwidth}
    \centering
    \includegraphics[width=\columnwidth]{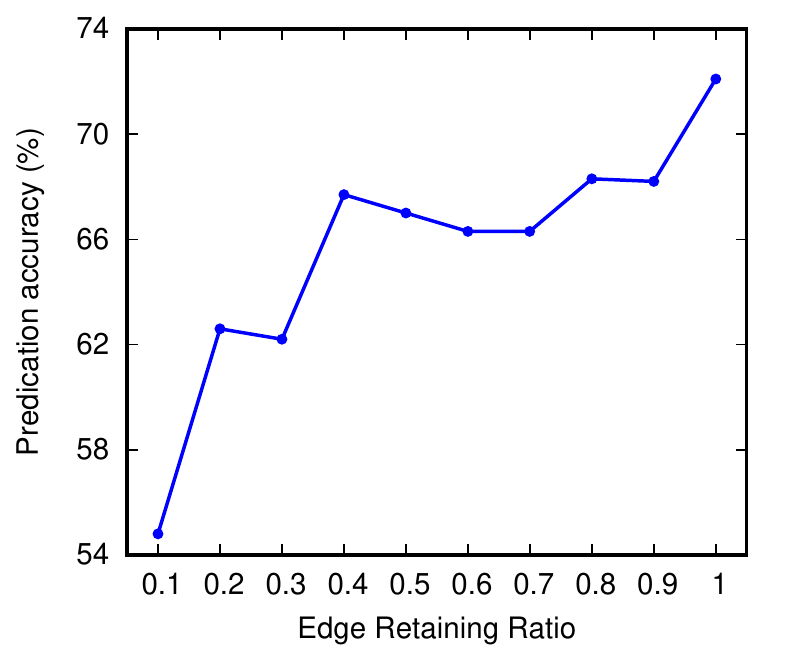}
    \caption{IMDB-BINARY}
    \end{subfigure}
    \hfill
    \begin{subfigure}[h]{0.49\columnwidth}
    \centering
    \includegraphics[width=\columnwidth]{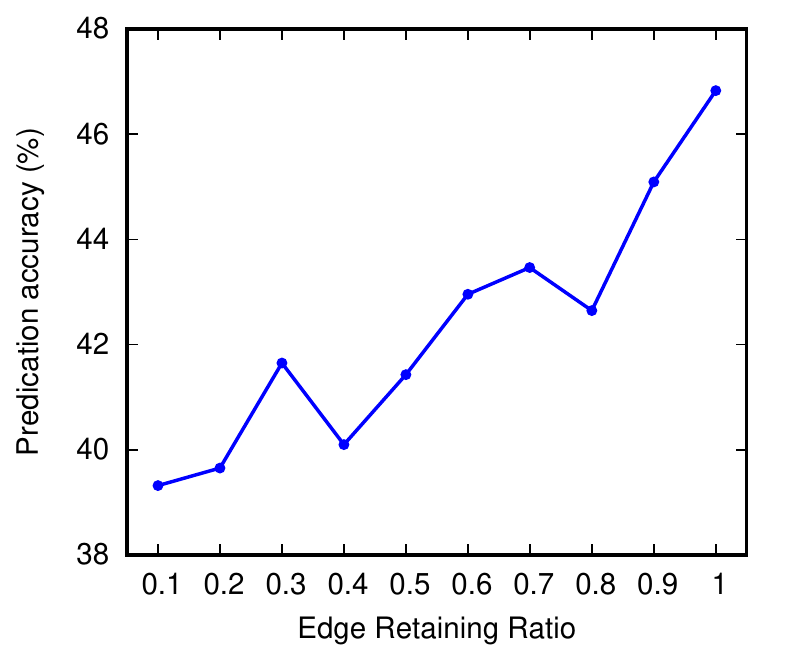}
    \caption{REDDIT-MULTI-12K}
    \end{subfigure}

    \caption{Accuracy change on different datasets with different edge retaining ratio ($\gamma$). Better view in colors.}
    \label{fig:diffedge}
\end{figure}

Fig.~\ref{fig:diffedge} plots the change in classification accuracy with respect to $\gamma$ on the six datasets. On the two labeled graphs (PROTEINS and D\&D), we find that keeping all edges ($\gamma = 1.0$) is not the best choice for graph classification. As shown in Fig.~\ref{fig:diffedge}(a) and (b), Co-Pooling achieves the highest classification accuracy when $\gamma = 0.7$ on PROTEIN, and $\gamma=0.6$ on D\&D, respectively. Similarly, this phenomenon can also be observed on the two attributed graphs (BZR-A and AIDS-A). As shown in Fig.~\ref{fig:diffedge}(c) and (d), Co-Pooling achieves the best performance when $\gamma$ is set to 0.6 on the two datasets. The results on the four datasets indicate not all edges are useful for graph classification when graphs have informative node attributes. Again, this confirms the benefit of our proposed method to preserve crucial edge information through edge-view pooling and use this knowledge to further guide node-view pooling.


In contrast, on the two plain graphs (IMDB-BINARY and REDDIT-MULTI-12K), keeping all edges renders the highest classification accuracy. As shown in Fig.~\ref{fig:diffedge} (e) and (f), Co-Pooling achieves the best performance on both graphs when retaining all edges ($\gamma=1.0$). This is what we have expected, because when graphs have no node attributes, the whole graph structure is more critical to graph classification.


\subsection{Graph Regression}
Lastly, we also carry out experiments to evaluate the effectiveness of the proposed method in undertaking graph regression task. We compare Co-Pooling with the same state-of-the-art pooling methods on the following two graph datasets:
\begin{itemize}
\item \textbf{ZINC}~\cite{bresson2019two,irwin2012zinc} contains 250k molecules and their property values. The task is to regress the property values of input graph. In this experiment, we focus on predicting one graph property, contained solubility. Following the setting in~\cite{dwivedi2020benchmarkgnns}, we use 10k graphs from ZINC for training, 1K graphs for validation, and 1K graphs for testing. 

\item \textbf{QM9}~\cite{wu2018moleculenet,ramakrishnan2014quantum} is a graph dataset consisting of 13k molecules with 19 regression targets. This dataset is originally used in quantum chemistry for regressing the property of molecules. We try to regress dipole moment $\mu$, one of 19 properties. All 13K molecules are randomly divided into 80\% training set, 10\% validation set and 10\% test set. 
\end{itemize}

For training a regression model for each dataset, we use GCN as the backbone GNN and inject two pooling layers before a MLP. Follow the setting in~\cite{dwivedi2020benchmarkgnns}, L1 loss function is used for training the model. We use Adam optimizer and learning rate decay policy to optimize the model. The initial learning rate and weight decay are set as 0.001 and 0.0001, respectively. We train the regression model under four different random seeds and report the average mean absolute error (MAE) on the test set. 

We compare our Co-Pooling method with SAGPool, ASAP, DiffPool, HGPSL, and EdgePool on the two datasets. As shown in Table~\ref{tab:regress}, our Co-Pooling method consistently achieves better regression performance than other baseline methods. Particularly, Co-Pooling outperforms DiffPool and HGPSL by a large margin on both datasets. The experimental results reflect the effectiveness of our method on graph regression task. This concludes that our method is effective in learning a better graph-level representation by fusing edge-view and node-view pooling, leading to competitive performance on both graph classification and regression tasks.


\begin{table}[htbp]
    \centering
    \caption{Graph regression task: mean absolute error (MAE) on ZINC dataset and QM9 dataset. Lower is better.}
    \label{tab:regress}
    \begin{tabular}{c|c|c}
    \toprule
        Datasets & ZINC & QM9 \\ 
        \midrule
        GCN+SAGPool & 0.378$\pm$0.031 & 0.545$\pm$0.010 \\
        GCN+ASAP & 0.372$\pm$0.026 & 0.500$\pm$0.017\\
        GCN+DiffPool & 1.641$\pm$0.026 & 1.331$\pm$0.014\\
        GCN+HGPSL &  1.326$\pm$0.096 & 1.035$\pm$0.049 \\
        GCN+EdgePool & 0.382$\pm$0.030 & 0.489$\pm$0.022\\
        \midrule
        GCN+Co-Pooling (ours) & \textbf{0.340$\pm$0.036} & \textbf{0.439$\pm$0.009} \\ 
         \bottomrule
        \end{tabular}
\end{table}

\section{Conclusion and Future Work}
We proposed a cross-view graph pooling (Co-Pooling) method to learn graph-level representations from both edge view and node view. We argued that most of the existing pooling methods are highly node-centric and they are unable to fully leverage crucial information contained in graph structure. To explicitly exploit graph structure, our proposed method seamlessly fuses the pooled graph information from two views. From the edge view, generalized PageRank is used to aggregate information from higher-hop neighbours to better capture the higher-order graph structure. The proximity weights between node pairs are calculated to prune less important edges. From the node view, the node importance scores are computed through the proximity matrix to select the top important nodes for node-view pooling. The pooled representations from two views are fused together as the final graph representation. Through cross-view interaction, edge-view pooling and node-view pooling complement each other to effectively learn informative graph-level representations. Experiments on a total of 15 graph benchmark datasets demonstrate the superior performance of our method on both graph classification and regression tasks. For future work, we would like to generalize cross-view graph pooling to learn more interpretable graph representations.


%

\appendices
\section{The graph neural network structure}
\label{app:GNN}
\begin{figure}[htbp]
    \centering
    \includegraphics[width=0.8\linewidth]{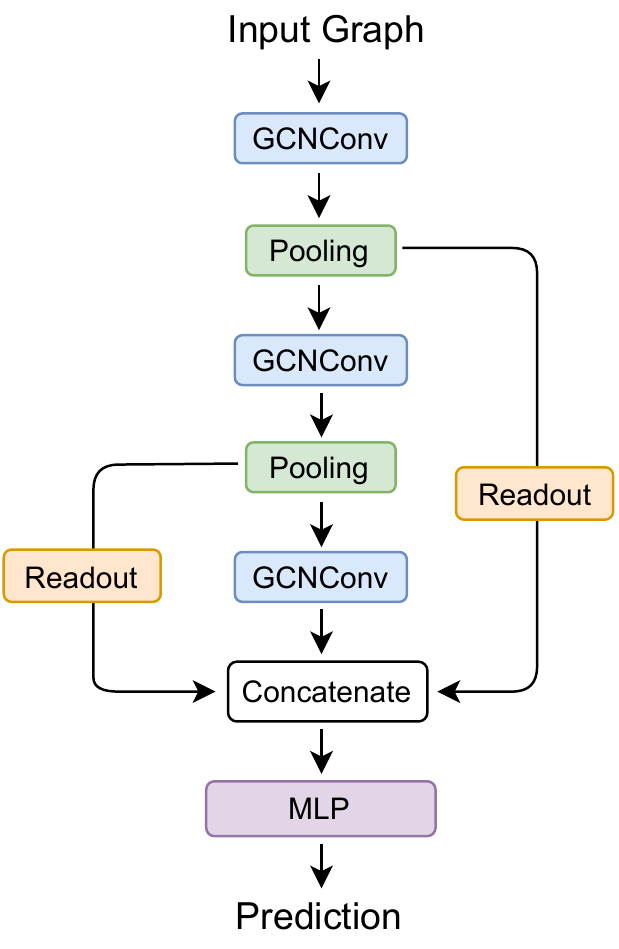}
    \caption{Graph neural network structure for graph classification task.}
    \label{fig:GNN}
\end{figure}

The network structure for graph classification task is shown in Fig.~\ref{fig:GNN}.

\ifCLASSOPTIONcaptionsoff
  \newpage
\fi



\bibliographystyle{IEEEtran}
\bibliography{reference.bib}
\end{document}